\documentclass{article} 
\usepackage[preprint]{colm2026_conference}

\usepackage{microtype}
\usepackage{hyperref}
\usepackage{url}
\usepackage{booktabs}
\usepackage{amsmath}
\usepackage{bm}
\usepackage{wrapfig}
\usepackage{multirow}
\usepackage{enumitem}
\usepackage{hhline}
\usepackage{booktabs}
\usepackage[table]{xcolor}
\usepackage[T1]{fontenc}

\usepackage{lineno}
\usepackage[most]{tcolorbox}
\definecolor{darkblue}{rgb}{0, 0, 0.5}
\hypersetup{colorlinks=true, citecolor=darkblue, linkcolor=darkblue, urlcolor=darkblue}
\definecolor{almond}{RGB}{239,222,205}
\definecolor{mossgreen}{RGB}{198,218,191}
\newcommand{\iapo}{IAPO}
\title{IAPO: Input Attribution-Aware Policy Optimization for \\Tool Use in Small Multimodal Agents}
\author{Yifan Yang, Zhen Zhang, Jiayi Tian, Liyan Tan, Zheng Zhang \\
University of California, Santa Barbara \\
\texttt{\{yifanyang,zhen\_zhang,jiayi\_tian,liyan\_tan,zzhang01\}@ucsb.edu}
}

\begin{document}

\ifcolmsubmission
\linenumbers
\fi

\maketitle

\begin{abstract}
This paper investigates reinforcement learning (RL) methods for improving tool-calling capabilities in multimodal small language model (SLM) agents. While existing works have explored various reward designs to improve agentic tool-calling ability, these approaches face inherent limitations for SLM training, especially under multimodal scenarios. First, many existing methods evaluate tool use correctness through exact matching against certain ground-truth or predefined formats. However, this assumption is often unsuitable for multimodal tasks, where multiple tool use paths may be valid and annotated tool trajectories are typically unavailable. Second, such sparse and brittle binary rewards provide little guidance on how to improve the underlying decision process, making them particularly difficult for multimodal SLM to learn from. To address these issues, we propose Input Attribution-Aware Policy Optimization (IAPO), an RL algorithm for improving tool use in multimodal SLM by aligning the model’s attribution\footnote{Attribution refers to the contribution of each input variable to model’s output \citep{deng2026attribution}.} across input components with that of a stronger teacher. Experiments on Qwen2.5-VL-3B show that the proposed method improves visual question answering accuracy by an average of 3\% across six test sets compared with existing visual tool use work, by helping the model attend to the most relevant input evidence.
\end{abstract}

\section{Introduction}
Since the development of methods such as Proximal Policy Optimization (PPO) \citep{schulman2017proximal} and Group Relative Policy Optimization (GRPO) \citep{shao2024deepseekmath}, reinforcement learning (RL) has emerged as a central paradigm for improving performance on complex long-horizon reasoning tasks \citep{xie2025logic, zhang2025survey}. Owing to its ability to directly optimize sequential decision making, RL has been widely used to train large language models (LLMs) as autonomous agents through online interaction with dynamic rollout trajectories \citep{zhang2026the}. 

At the same time, the increasing scale of LLMs has made inference increasingly expensive, posing a major obstacle to their practical deployment as agents and thereby motivating growing interest in small language model (SLM) agents. For instance, \citet{kang2025distilling} propose an agent distillation framework based on supervised fine-tuning (SFT), where a student model learns to imitate the behavior of a stronger teacher agent. More recently, \citet{lyu2026mock} introduce a new data collection pipeline for synthesizing training data for RL-based SLM agent training, building on the observation that RL can provide stronger training signals than SFT in agent settings \citep{mai2025agent}.

Despite these advances, agentic RL training with the vanilla GRPO algorithm still yields suboptimal performance. To address this issue, several recent studies have explored more specialized RL algorithms with fine-grained reward mechanisms that provide more informative feedback for tool use in general LLM agents. For instance, ToRL introduces a reward formulation that combines answer correctness with execution-based penalties to better supervise tool-calling behavior \citep{li2025torl}. Similarly, TRM proposes a dedicated process reward model to assess tool invocation quality, thereby enabling explicit verification of tool-call correctness \citep{ma2026empowering}.

However, directly adapting these reward designs does not yield optimal performance for multimodal SLM training, mainly for two reasons. First, existing approaches rely on rigid format checks or exact matching against ground-truth tool trajectories, while multimodal tasks often admit multiple valid tool use paths. For example, when focusing on relevant chart regions, a model may highlight, mask, or draw bounding boxes—different actions that all provide useful visual evidence for reasoning. Second, such binary, match-based rewards are often sparse and brittle, providing limited guidance for SLMs to learn the decision-making process. 

Furthermore, these methods focus exclusively on the textual domain. In the relatively limited line of work on multimodal tool use, approaches such as VTool-R1 \citep{wu2026vtoolr} and OpenThinkIMG \citep{su2025openthinkimg} still rely on the vanilla GRPO algorithm. However, as we show in Section~\ref{sec:motiviting}, multimodal SLMs trained with vanilla GRPO continue to produce a substantial number of misused tool calls even in the later stages of training. This indicates that RL algorithm design for multimodal tool use remains underexplored, particularly for small-scale models.

To provide richer guidance for multimodal SLMs during RL training, a natural idea is to leverage a stronger teacher model to provide supervision over the student's intermediate decision process, inspired by knowledge distillation \citep{gou2021knowledge}. The key challenge, however, is that RL rewards are typically computed from decoded text output, making it difficult to capture the model's internal knowledge. Fortunately, we find that the attribution of the input to the generated tool call is strongly correlated with tool-call correctness: when the appropriate parts (e.g. definition for correct tool) of the input contribute more to the model's output, it is substantially more likely to select the correct tool. This observation creates a bridge for designing rewards that reflect the model's internal knowledge, by using input attribution as a dense supervision signal for online RL.

Building on this insight, we propose \textbf{Input Attribution-Aware Policy Optimization} (\iapo), the first method to design a reward specific for SLM training. Specifically, we quantify the contribution of different input components to the generated tool calls, and use the KL divergence between the student and teacher attribution distributions to derive a dense, distillation-style guidance signal. This reward not only improves tool selection, but also provides explicit supervision over the underlying decision-making process. Our key contributions are as follows:
\begin{itemize}
\item We propose the new integrated gradients (IG) scores, which quantify the contributions of different input components to the tool calls.
\item To transfer the teacher's tool-calling capability to the SLM during RL training, we propose \iapo, which applies an attribution-based penalty by comparing the IG scores of the student and teacher.
\item We conduct case studies showing that multimodal SLMs still suffer from tool misuse even after training with vanilla GRPO. Furthermore, we show that directly adapting tool-use reward designs developed for large text-only models is inadequate for achieving optimal performance in multimodal SLM training, as \iapo\ improves accuracy by more than 3\% over these adaptations.
\end{itemize}
\section{Related Works} 
In this paper, we consider a multimodal tool-calling setting in which models are prompted to use visual editing tools to improve reasoning performance on visual question answering tasks. While prior work has sought to improve text-only reasoning conditioned on multimodal inputs \citep{shen2025vlm, huang2025vision}, multimodal learning with genuine image-involved tool use has received comparatively little attention until recently. Early inference-time methods introduced intermediate visual reasoning steps to enhance multimodal reasoning. For instance, Visual Sketchpad \citep{hu2024visual} and Refocus \citep{fu2025refocus} allow models to generate visual artifacts or invoke image editing tools, such as cropping, highlighting, and masking, during reasoning, thereby enabling iterative focus on task-relevant visual regions.

More recently, RL has been adopted to train multimodal agents that learn when and how to invoke tools during reasoning. For example, DeepEyes encourage VLMs to ``think with images'' by incorporating image-involved tool calls into RL training \citep{zheng2025deepeyes}. Along the same line, works like OpenThinkIMG \cite{su2025openthinkimg} and VTool-R1 \citep{wu2026vtoolr} integrates visual editing tools into an RL fine-tuning framework, allowing VLMs to interleave textual reasoning with intermediate visual operations that improve final task accuracy. These works collectively demonstrate the potential of RL for improving tool-calling performance in VLMs. However, the design of fine-grained intermediate rewards for guiding multimodal tool selection and reasoning remains largely unexplored.
\begin{figure}
    \centering
    \includegraphics[width=0.9\linewidth]{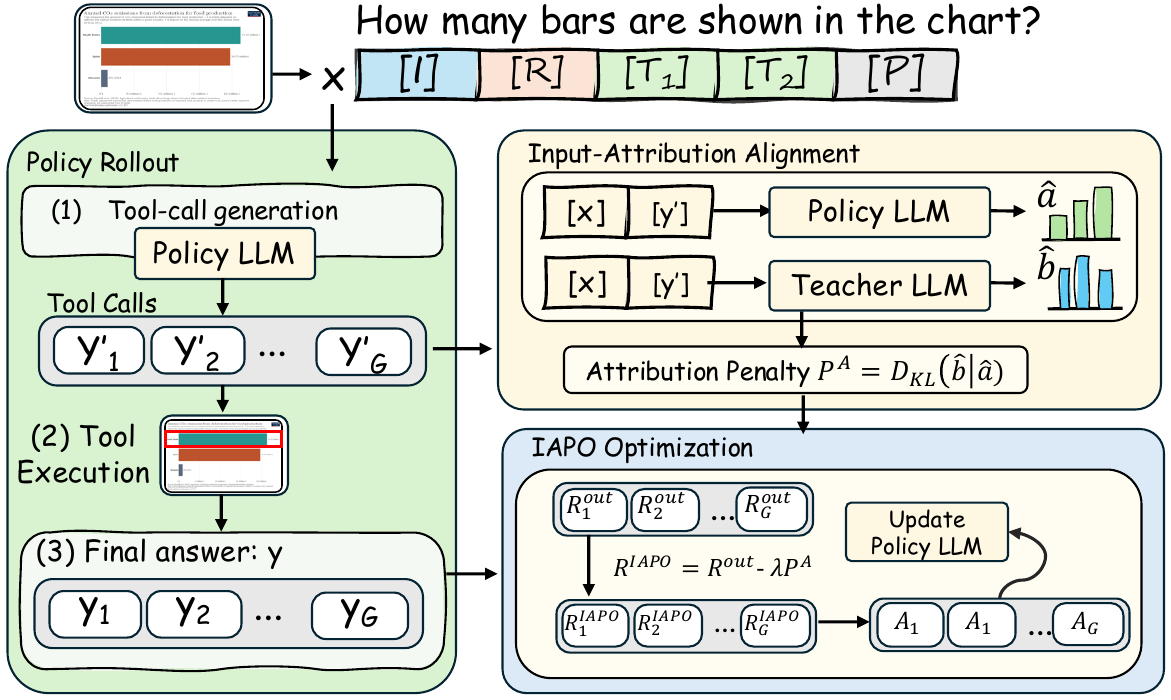}
    \caption{Overview of the IAPO method. The attribution penalty \(P^A\) is computed from the tool-call sequence \(\mathbf{y}'\) generated in the first rollout. The final IAPO reward combines the standard outcome-based reward \(R^{\mathrm{out}}\) with the attribution penalty \(P^A\).}
    \label{fig:placeholder}
\end{figure}
\section{Methods}
\subsection{Training VLM Agents with Vanilla GRPO}
In this section, we start by presenting the problem formulation under the vanilla GRPO setting and analyze the limitations of vanilla GRPO reward design for visual tool-calling tasks. We follow the vanilla-GRPO-based training framework for visual tool-calling agents introduced in VTool-R1 \citep{wu2026vtoolr}. Concretely, we consider a VLM policy $\pi_{\theta}$, parameterized by $\theta$, for multimodal tool-calling. In each rollout, the policy takes an image $\bm{I}$ and a prompt $\bm{x}$ as input, and generates an output response $\bm{y}$. 

\begin{figure*}[t]
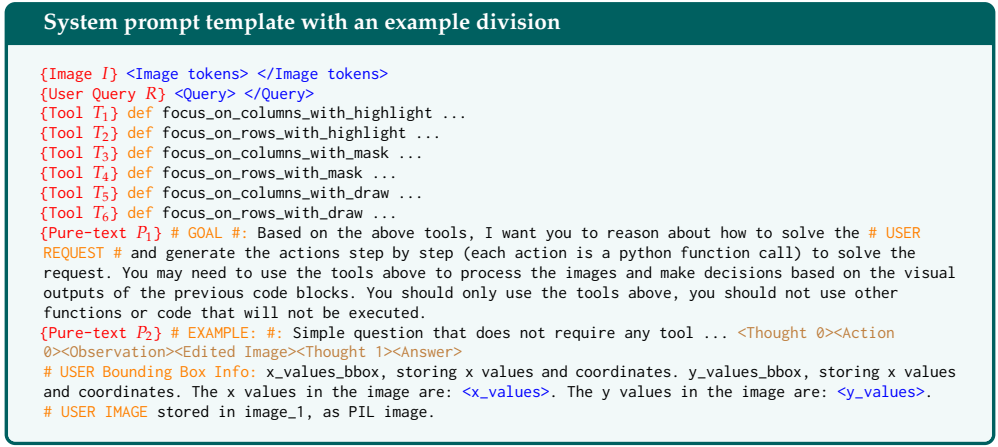

\centering
\resizebox{0.95\textwidth}{!}{
\begin{tcolorbox}[
    halign=flush left,
    breakable, 
    colback=teal!5!white,
    colframe=teal!75!black,
    title={\small\textbf{System prompt template with an example division}},
    fonttitle=\bfseries,
    fontupper=\ttfamily,
    width=\columnwidth,  
    enhanced jigsaw,
]
\scriptsize 
{\color{red} \{Image $I$\}} {\color{blue} <Image tokens>} {\color{blue} </Image tokens>}

{\color{red} \{User Query $R$\}} {\color{blue} <Query>} {\color{blue} </Query>}

{\color{red} \{Tool $T_1$\}} {\color{orange} def} \texttt{focus\_on\_columns\_with\_highlight} ...

{\color{red} \{Tool $T_2$\}} {\color{orange} def} \texttt{focus\_on\_rows\_with\_highlight} ...

{\color{red} \{Tool $T_3$\}} {\color{orange} def} \texttt{focus\_on\_columns\_with\_mask} ...

{\color{red} \{Tool $T_4$\}} {\color{orange} def} \texttt{focus\_on\_rows\_with\_mask} ...

{\color{red} \{Tool $T_5$\}} {\color{orange} def} \texttt{focus\_on\_columns\_with\_draw} ...

{\color{red} \{Tool $T_6$\}} {\color{orange} def} \texttt{focus\_on\_rows\_with\_draw} ...

{\color{red} \{Pure-text $P_1$\}} {\color{orange}\# GOAL \#:} Based on the above tools, I want you to reason about how to solve the {\color{orange}\# USER REQUEST \#} and generate the actions step by step (each action is a python function call) to solve the request.
You may need to use the tools above to process the images and make decisions based on the visual outputs of the previous code blocks.
You should only use the tools above, you should not use other functions or code that will not be executed.

{\color{red} \{Pure-text $P_2$\}} {\color{orange}\# EXAMPLE: \#:} Simple question that does not require any tool ...
{\color{brown} <Thought 0><Action 0><Observation><Edited Image><Thought 1><Answer>}
\\ {\color{orange}\# USER Bounding Box Info:} x\_values\_bbox, storing x values and coordinates. y\_values\_bbox, storing x values and coordinates. The x values in the image are:  {\color{blue}<x\_values>}. The y values in the image are:  {\color{blue}<y\_values>}.
\\ {\color{orange}\# USER IMAGE} stored in image\_1, as PIL image. 
\end{tcolorbox}
}
\caption{System prompt template with an example division used in IAPO rollouts.}
\label{fig:prompt_abbrev}
\end{figure*}

As illustrated in Fig.~\ref{fig:prompt_abbrev}, we divide the input prompt \(\bm{x}\) into an ordered sequence of semantic blocks $\bm{x} = [\bm{b}_1, \bm{b}_2, \dots, \bm{b}_K]$, where each block corresponds to one functional part of the prompt. In this work, we consider four block types: pure-text context (\(P\)), user query (\(R\)), image placeholder (\(I\)), and tool definition (\(T\)). We use \(\tau(k) \in \{P, R, I, T\}\) to denote the type of block \(\bm{b}_k\). For example, a prompt may be written as $\bm{x} = [P_1, T_1, T_2, P_2, R, I]$, where the order follows the actual prompt structure. Since the system prompt template is fixed across all samples, we use the same block partition throughout the paper.

For the rollout process, we adopt a two-turn setup. In the first turn, the model is prompted with input $\bm{x}$ and image $\bm{I}$ to generate tool calls $\bm{y}’ \sim \pi_{\theta}(\cdot \mid \bm{x}, \bm{I})$, where the original image $\bm{I}$ is processed by code executor $\mathcal{T}$ to obtain the intermediate edited image $\bm{I}’ = \mathcal{T}(\bm{y}’, \bm{I})$. In the second turn, the model performs reasoning over the modified visual input to produce the final answer $\bm{y} \sim \pi_{\theta}(\cdot \mid \bm{x}, \bm{y}’, \bm{I}, \bm{I}’)$. We do not consider settings with more than two turns due to the limited availability of VLM reasoning datasets suitable for sequential tool use. In addition, the long context window introduced by the large number of intermediate image tokens would exceed our computational budget.

Existing visual tool-reasoning methods based on vanilla GRPO, such as VTool-R1 and OpenThinkIMG, typically use a binary outcome reward \(R_i^{\mathrm{out}}\), which assigns \(1\) if the final answer in the \(i\)-th response is correct and \(0\) otherwise. For each multimodal input sample \((\bm{I}, \bm{x})\sim\mathcal{D}\), GRPO samples a group of \(G\) responses \(\{\bm{y}_1, \dots,\bm{y}_i,\cdots, \bm{y}_G\}\) from the old policy \(\pi_{\mathrm{old}}(\cdot \mid \bm{I}, \bm{x})\). The GRPO objective is then defined as
\begin{align*}
&\mathcal{J}_{\mathrm{GRPO}}(\theta)=\mathbb{E}_{\mathcal{D},\, \{\bm{y}_i\}_{i=1}^{G}\sim \pi_{\mathrm{old}}(\cdot \mid \bm{I},\bm{x})}\\
&
\Bigg[
\frac{1}{G}
\sum_{i=1}^{G}
\frac{1}{|\bm{y}_i|}
\sum_{t=1}^{|\bm{y}_i|}
\min\Big(
r_{i,t}(\theta)\hat{A}_i,\,
\operatorname{clip}(r_{i,t}(\theta), 1-\epsilon, 1+\epsilon)\hat{A}_i
\Big)
\notag
-\beta D_{\mathrm{KL}}\!\left[\pi_\theta \,\|\, \pi_{\mathrm{ref}}\right]
\Bigg],
\end{align*}
where the scalar \(\epsilon\) is the clipping threshold in GRPO, \(\beta\) is the coefficient for KL regularization, and \(\pi_{\mathrm{ref}}\) denotes the reference policy. The token-level importance ratio \(r_{i,t}(\theta)\) is defined as $r_{i,t}(\theta) = \frac{\pi_{\theta}(y_{i,t}\mid \bm{x}, y_{i,<t})}{\pi_{\mathrm{old}}(y_{i,t}\mid \bm{x}, y_{i,<t})}$  for the \(t\)-th token in the response \(\bm{y}_i\), and \(y_{i,<t}\) denotes the prefix consisting of all tokens before position \(t\). Since the reward is defined at the response level, the same advantage value is shared across all tokens in the same response. Specifically, the group-relative normalized advantage \(\hat{A}_i\) for the \(i\)-th response is computed as
\begin{align*}
\hat{A}_i
=
\frac{
R_i^{\mathrm{out}} - \mathrm{mean}(\{R_j^{\mathrm{out}}\}_{j=1}^{G})
}{
\mathrm{std}(\{R_j^{\mathrm{out}}\}_{j=1}^{G})
},
\end{align*}
where \(\mathrm{mean}(\{R_j^{\mathrm{out}}\}_{j=1}^{G})\) and \(\mathrm{std}(\{R_j^{\mathrm{out}}\}_{j=1}^{G})\) denote the mean and standard deviation of the outcome rewards across the sampled group, respectively.

Although GRPO can improve overall task performance, its vanilla objective optimizes only the correctness of the final output and provides no direct supervision over the quality of intermediate tool use decisions. This creates a severe credit assignment problem for multimodal tool calling: the model may invoke an unnecessary or incorrect tool, attend to irrelevant visual regions, or rely on spurious textual cues, yet still receive a high reward as long as the final answer happens to be correct. 

This issue is especially problematic for SLMs, whose weaker reasoning and grounding capabilities make them less able to infer correct tool use behavior from sparse outcome-level rewards alone. As a result, SLMs are more likely to overfit to superficial correlations in the prompt or exploit accidental shortcuts, rather than learning a robust tool-selection policy grounded in the relevant multimodal evidence. This limitation is validated by the experimental results in Section~\ref{sec:motiviting}, where the model still suffers from substantial tool misuse even after extensive GRPO training. Therefore, vanilla GRPO cannot reliably distinguish genuinely effective tool use behavior from accidental success, particularly for smaller multimodal agents.
\subsection{\iapo: Input Attribution-Aware Policy Optimization}
In this section, we propose \iapo, an RL algorithm for multimodal SLM training that introduces a new reward design combining the final outcome reward \(R^{\mathrm{out}}\) with an attribution-based penalty \(P^{A}\). This penalty distills guidance from a stronger teacher model by encouraging the SLM to focus on the input components that are most relevant for correct tool-use decisions. We first introduce a new integrated gradients (IG) score that forms the core of the penalty, and then present the full \iapo\ reward formulation.

\textbf{Integrated Gradient Score:} Intuitively, the IG score uses gradient-based attribution to measure how different semantic blocks in the input prompt \(\bm{x}\) influence the model's tool-calling behavior. In particular, it captures both implementation invariance and input sensitivity, making it well suited for quantifying the contribution of each input component to the predicted tool call. For each input token \(i\), the IG score is computed by combining its input embedding \(\bm{e}_i\) with the corresponding gradient \(\bm{g}_i\) of the tool-call score with respect to that embedding. Formally, the IG score is defined as follows.

Following the prompt formulation above, we represent the textual input as an ordered sequence of $K$ semantic blocks, $\bm{x} = [\bm{b}_1, \bm{b}_2, \dots, \bm{b}_K]$, where each block $\bm{b}_k$ is assigned a semantic type by the mapping $\tau(k) \in \{P,R,I,T\}$. After combining the textual prompt $\bm{x}$ with the input image $\bm{I}$ and applying the multimodal tokenizer and processor, the VLM produces a merged input embedding matrix $\bm{E} = [\bm{e}_1, \dots, \bm{e}_S] \in \mathbb{R}^{S \times d},$ where $S$ is the total input length after multimodal tokenization and $d$ is the hidden dimension. Let $\Omega_k \subseteq \{1,\dots,S\}$ denote the set of token indices corresponding to block $\bm{b}_k$ in the merged input sequence.

For a given rollout, let $\bm{y}' = [y'_1,\dots,y'_{L'}]$ denote the generated tool-call sequence, where $L'$ is the sequence length. To measure how the input supports this tool call, we first define a scalar attribution target as the log-probability of the full tool-call sequence under the current policy:
\[
s(\bm{E}) = \log \pi_\theta(\bm{y}' \mid \bm{E})
= \sum_{t=1}^{L'} \log \pi_\theta(y'_t \mid y'_{<t}, \bm{E}).
\]
In practice, this score is computed by first obtaining the model logits over the vocabulary at each decoding step, converting them into log-probabilities, and then selecting the log-probability of each generated tool-call token. Summing these token-level log-probabilities gives the final sequence-level score \(s(\bm{E})\).

We then compute the gradient of this score with respect to the input embedding for the $i$-th token:
\[
\bm{g}_i = \frac{\partial s(\bm{E})}{\partial \bm{e}_i}, \qquad i=1,\dots,S.
\]
Here, \(\bm{g}_i \in \mathbb{R}^d\) measures how sensitive the tool-call score is to the embedding of the \(i\)-th input token.

Based on this gradient, we define the token-level attribution using a gradient-times-embedding formulation:
\[
o_i = \left\| \bm{g}_i \odot \bm{e}_i \right\|_2,
\]
where \(\odot\) denotes element-wise multiplication. Intuitively, \(o_i\) measures how much the \(i\)-th input token contributes to increasing the probability of the generated tool-call sequence, while also taking into account the scale of its embedding representation.

Finally, we aggregate token-level attribution within each semantic block $O_k = \sum_{i\in\Omega_k} o_i$. The resulting block-level score $O_k$ provides an input-grounded attribution signal indicating which prompt components most strongly influence the model's tool-calling decision.

\paragraph{Input Attribution-Aware Policy Optimization (\iapo)}
As we mentioned, IAPO extend GRPO by augmenting the trajectory-level outcome reward with an attribution penalty term. Let $\mathbf{a}_i = \{O^a_1, \cdots, O^a_K\}$ and $\mathbf{b}_i= \{O^b_1, \cdots, O^b_K\}$ denote the actor and teacher IG scores over the $K$ input parts. After normalizing them into distributions $\hat{\mathbf{a}}_i$ and $\hat{\mathbf{b}}_i$, we define the attribution alignment penalty as $P_i^{A} = D_{\mathrm{KL}}(\hat{\mathbf{b}}_i \,\|\, \hat{\mathbf{a}}_i)$. We then define the modified trajectory reward as
\begin{align}\label{eq:reward}
    R_i^{\mathrm{\iapo}} = R_i^{out} - \lambda P_i^{A},
\end{align}
where $\lambda$ controls the strength of the attribution alignment regularization and $R^{out}$ denote outcome reward.

Given a group of $G$ sampled responses associated with the same input, we compute the group-relative advantage as
\[
A_i^{\mathrm{\iapo}} =
\frac{
R_i^{\mathrm{\iapo}} - \mathrm{mean}(\{R_j^{\mathrm{\iapo}}\}_{j=1}^{G})
}{
\mathrm{std}(\{R_j^{\mathrm{\iapo}}\}_{j=1}^{G}) + \epsilon
}.
\]

Following standard GRPO, we broadcast $A_i^{\mathrm{\iapo}}$ to all response tokens and use the resulting token-level advantage $A_{i,t}^{\mathrm{\iapo}}$ in the clipped surrogate objective. In this way, the policy is encouraged not only to maximize the original task reward, but also to align its tool-calling behavior with the teacher-provided input attribution patterns.

\section{Experiments}
In this section, we present the experimental results to show the effectiveness of our IAPO method. We first present a case study to illustrate the motivation and intuition behind our method, followed by comprehensive experiments comparing against existing baselines. Moreover, we include an ablation study analyzing the effects of hyperparameter choices and teacher model selection in Appendix \ref{sec:ablation}.

\subsection{Training Details}\label{sec:setup}
We evaluate \iapo\ on visual question answering tasks with visual editing tools using the Qwen2.5-VL model family \citep{bai2025qwen25vltechnicalreport}. We conduct RL training with the VeRL framework \citep{sheng2025hybridflow} and the AdamW optimizer \citep{loshchilov2019decoupled}. For \iapo\ training on Qwen2.5-VL-3B-Instruct, we first train a Qwen2.5-VL-7B-Instruct teacher model using standard GRPO on the same training set. The \iapo\ pipeline contains two stages: (1) a short cold-start GRPO phase that equips the model with basic tool-calling capability, and (2) the main \iapo\ training phase using the reward defined in Eq.~(\ref{eq:reward}). Unless otherwise noted, all experiments use Qwen2.5-VL-3B-Instruct as the backbone and are conducted on NVIDIA Tesla A100-40GB GPUs. Details of the training and testing datasets can be found in Appendix~\ref{app:dataset}.

\textbf{Baselines:} We consider two categories of baselines: existing multimodal tool-calling agents and prior methods for intermediate reward design in tool-calling RL. The baselines are summarized below:
\begin{itemize}[leftmargin=*, nosep]
    \item \textbf{Off-the-shelf Models:} We report inference-only results for GPT-4o \citep{openai2024gpt4technicalreport} and Qwen2.5-VL-3B models.
    \item \textbf{VTool-R1:} A representative multimodal tool-calling baseline trained with vanilla GRPO.
    \item \textbf{TRM:} TRM introduces an additional reward model to verify the correctness of tool calls, making it a relevant baseline for tool-calling reward design. For a fair and convenient comparison, we reimplement TRM in our visual setting by using GPT-4o as the verifier.
    \item \textbf{ToRL:} ToRL designs a tool-calling reward based on both final-task accuracy and tool-execution penalties. Following its core idea, we incorporate the same $-0.5$ penalty for failed tool executions as a comparison baseline.
\end{itemize}

\textbf{Tool set:} We follow VTool-R1 for the design of tool set for a fair compreison, which contains a total of 6 visual editing tools across three types (mask out, draw box and highlight color) with Python code. The definitions of these tools are as follows:
\begin{itemize}[leftmargin=*, nosep]
    \item \texttt{focus\_on\_columns\_with\_highlight}: highlights relevant columns with a light red overlay.
    \item \texttt{focus\_on\_rows\_with\_highlight}: highlights relevant rows with a light red overlay.
    \item \texttt{focus\_on\_columns\_with\_mask}: masks irrelevant columns with a white overlay.
    \item \texttt{focus\_on\_rows\_with\_mask}: masks irrelevant rows with a white overlay.
    \item \texttt{focus\_on\_columns\_with\_draw}: marks relevant columns with a solid red bounding box.
    \item \texttt{focus\_on\_rows\_with\_draw}: marks relevant rows with a solid red bounding box.
\end{itemize}
\begin{figure}[t]
    \centering
    \includegraphics[width=0.9\linewidth]{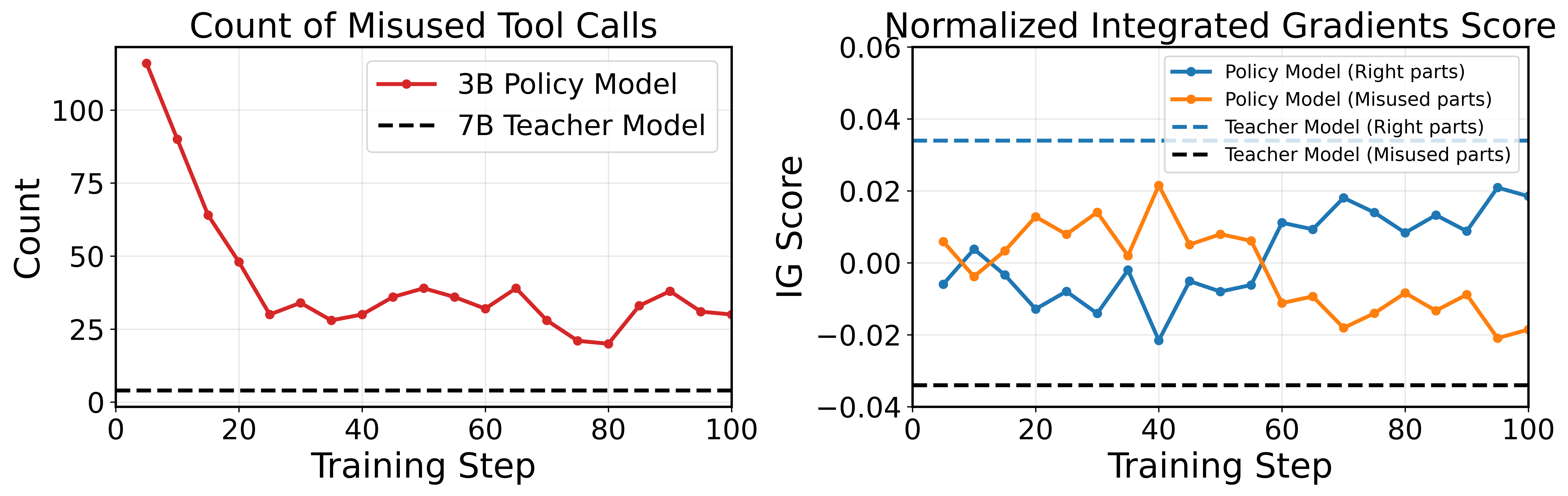}
    \caption{Case study: Count for misused tool calls and IG Scores with selected dataset. We use Qwen2.5-VL-3B as the policy model and a well-trained Qwen2.5-VL-7B model as the teacher. Both models are trained with the vanilla GRPO method used in VTool-R1.}
    \label{fig:case}
\end{figure}
\subsection{Case Study on the Intuition Behind \iapo} \label{sec:motiviting}
To build intuition for \iapo, we conduct a case study that analyzes both tool-calling behavior and input attribution patterns during training under a simplified experimental setup. This study is designed to highlight the limitations of vanilla GRPO training and the utility of IG scores in diagnosing tool-selection errors. Specifically, we examine how the model’s tool usage evolves over time and whether IG score can reveal failure modes in tool calling.

To make the optimal tool choice clear, we select 256 samples from the ChartQA validation set that are clearly \textit{column-divisible} or \textit{row-divisible}, meaning that the chart contains well-separated vertical or horizontal groups, bars, or regions such that column-based or row-based tools can be applied naturally and reliably. This construction allows us to determine which type of tool should be selected from the tool set introduced in Section~\ref{sec:setup} clearly.

To analyze tool-calling behavior and IG attribution during training, we evaluate the selected validation subset every 5 training steps throughout vanilla GRPO training (the same training scheme used in VTool-R1) on a 3B model. At each checkpoint, we record both the total number of misused tool calls and the normalized IG scores assigned to the right and misused part of tool definitions. Specifically, for column-divisible plots, column-based tools are regarded as correct and row-based tools as incorrect, while the opposite holds for row-divisible plots. The results are shown in Fig.~\ref{fig:case}. Specifically, we report the total number of incorrect tool calls across training steps, where an incorrect call corresponds to applying row-based tools to column-divisible charts or column-based tools to row-divisible charts. We also compute the mean IG score over the prompt tokens associated with different tool groups. As a reference point, we further evaluate a well-trained 7B teacher model under the same protocol and report its tool-misuse count and IG attribution statistics. Based on these results, we draw the following observations:

\textbf{SLM still make tool-selection errors after GRPO training.}
Even after 100 steps of GRPO training, the SLM still makes 30 misused tool calls out of 256 evaluated samples. This occurs because an incorrect tool call does not necessarily prevent the model from reaching the correct final answer; in some cases, the model can recover through its own reasoning ability. As a result, outcome-based RL training may fail to adequately penalize tool-selection errors. This issue is even more pronounced in the early stage of training, where nearly half of the tool calls are incorrect. These results suggest that tool-selection accuracy remains a key challenge for smaller multimodal LLMs.

\textbf{IG scores reflect tool-call correctness.}
By examining the attribution curves for different tool groups, we observe that the IG score assigned to the correct input parts gradually increases during training and eventually becomes consistently higher than that assigned to the misused tool definitions. This trend aligns closely with the decrease in the number of incorrect tool calls. The result suggests that tool misuse is strongly correlated with misallocated attention over the prompt, and that IG scores provide a meaningful signal for diagnosing tool-selection failures.

\textbf{Larger models exhibit stronger tool-selection ability.}
We further observe that the 7B teacher model achieves better tool-selection accuracy with only 4 incorrect tool calls and shows a clearer separation between the IG scores assigned to correct versus incorrect input parts. This indicates that larger models not only select tools more reliably, but also exhibit more appropriate attribution patterns over the input. These findings support the use of a stronger teacher model to guide the student policy during training.
\setlength{\tabcolsep}{8pt}
\begin{table*}[t]
\centering
\vspace{-1em}
{\fontsize{9pt}{13pt}\selectfont
\resizebox{0.9 \textwidth}{!}{%
\begin{tabular}{l|ccccccc}
\toprule[1.2pt]
& \multicolumn{3}{c}{\textbf{Chart}} & \multicolumn{3}{c}{\textbf{Table}} & \multirow{2}{*}{Avg}\\
\cmidrule(lr){2-4} \cmidrule(lr){5-7}
Methods & CharXiv & Horizontal & Vertical & VWTQ & VWTQ\_syn & VTabFact \\
\hline
\rowcolor{almond}
\multicolumn{8}{c}{\textit{Large Scale Multimodal LLMs}} \\
GPT-4o & 47.9 & 75.7 & 81.7 & 67.5 & 72.8 & 91.6 & 72.9 \\
Qwen-2.5VL-7B &31.0   & 69.6 & 48.7 & 37.5 & 52.6 & 45.6 & 47.5 \\
+ VTool-R1 (GRPO) & 39.2 & 81.3 & 75.4 & 52.6 & 68.4 & 69.1 & 64.3 \\
\hline
\rowcolor{mossgreen}
\multicolumn{8}{c}{\textit{Small Scale Multimodal LLMs}} \\
Qwen-2.5VL-3B & 18.9    & 25.7           & 22.8         & 14.2 & 26.3      & 36.8     & 24.1 \\
+ VTool-R1 (GRPO) & 23.1    & 63.7           & 60.9         & 46.6 & 52.6      & 70.5     & 52.9 \\
+ TRM* & 23.6    & 64.3           & 62.7         & 44.3 & 54.6      & 70.9     & 53.4 \\
+ ToRL* & 24.1    & 64.0           & 63.1         & 44.4	&56.1	&71.0	&53.8  \\
+ \textsc{\iapo} & 26.1    & 66.0           & 64.1         &46.7 &	60.2 &	72.1 &	55.9  \\
\bottomrule[1.2pt]
\end{tabular}%
}
}
\vspace{-1ex}
\caption{Performance comparison of baseline methods. * Since TRM and ToRL were originally proposed for text-only tool-use settings, we adapt them to visual tool-use setup. For TRM, we directly use GPT-4o as the tool-call reward model for convenience, instead of training a separate offline model on GPT-4o-generated data as done in original paper.}
\label{tab:baselines}
\end{table*}
\subsection{Comparison with Baselines}
In this section, we compare our \iapo\ method with the baseline methods introduced in Section~\ref{sec:setup} on chart and table question-answering tasks. Specifically, for each method, we train two separate models for the chart and table settings, and report test accuracy on six evaluation sets in Table~\ref{tab:baselines}. From the results, we draw the following conclusions:

\textbf{Small MLLMs struggle to use tools effectively.} As shown in Table~\ref{tab:baselines}, the off-the-shelf Qwen-2.5VL-3B model has limited ability to solve these tasks through tool use. Simply prompting the model to call tools at inference time does not lead to strong performance. In contrast, larger models such as Qwen-2.5VL-7B and GPT-4o achieve substantially better results under inference-only tool use, suggesting that effective tool use remains challenging for small multimodal models.

\textbf{Existing reward designs remain suboptimal.} Recent reward-design methods, such as TRM and ToolRL, improve RL-based visual tool calling over vanilla GRPO. However, they still consistently underperform \iapo. Notably, \iapo\ achieves stronger performance than TRM even when paired with a substantially smaller teacher model. This result suggests that deterministic verification of tool-call correctness is not well suited to visual tool-calling tasks, where multiple tool choices may be valid for solving the same problem.

\textbf{IAPO exhibits stronger generalization ability.} As shown by the results on the out-of-distribution CharXiv benchmark, IAPO achieves substantially better performance than existing methods. This suggests that IAPO generalizes better beyond the training distribution. We attribute this advantage to the transfer of input-grounding behavior from the teacher model, which helps the policy focus on more relevant evidence and reduces overfitting during RL training.
\subsubsection{Training Curve and Detailed Analysis}
\begin{figure}[t]
    \centering
\includegraphics[width=0.85\linewidth]{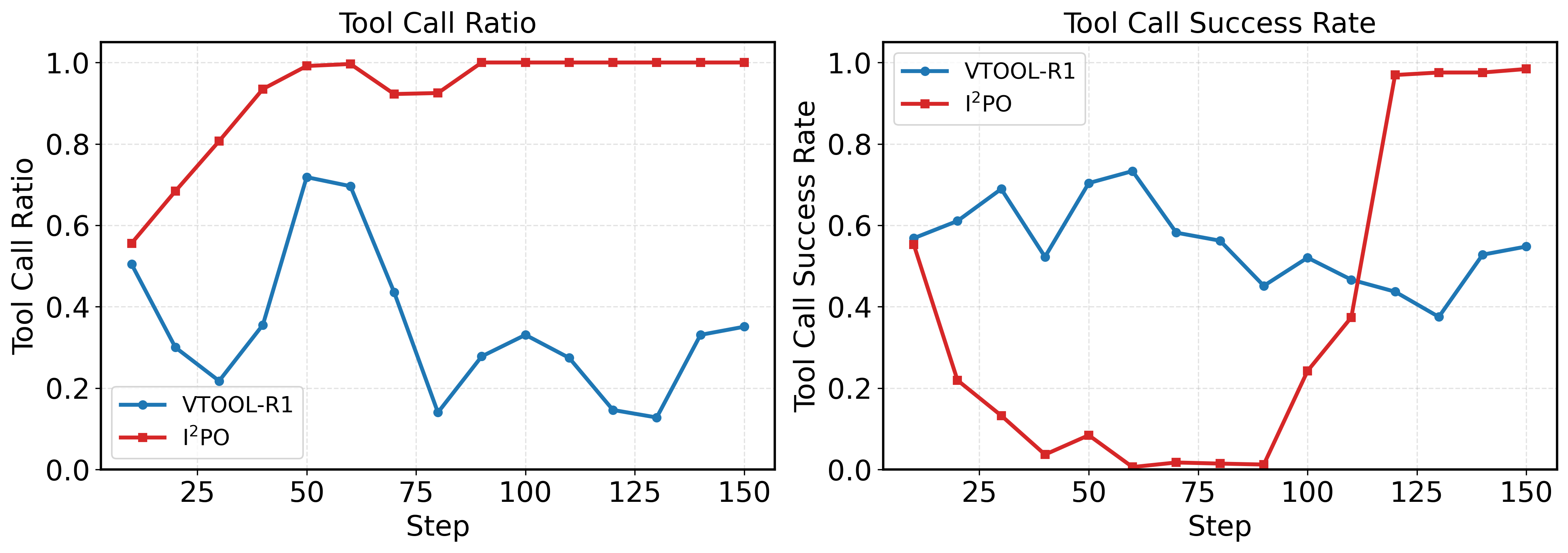}
    \caption{Tool use behaviors during training.}
    \label{fig:tool_behavior}
\end{figure}
In this section, we further analyze the training dynamics of \iapo\ by comparing it with the vanilla GRPO baseline, VTool-R1, using training-time tool use statistics. Due to limited computational resources, we report validation results over the first 150 training steps. As shown in Fig.~\ref{fig:tool_behavior}, \iapo\ exhibits substantially stronger tool use behavior than VTool-R1 throughout training. In particular, \iapo\ maintains a much higher tool-call ratio, quickly increasing to nearly 1.0 and remaining stable thereafter, whereas VTool-R1 shows a substantially lower and more unstable tendency to invoke tools. This indicates that \iapo\ more effectively encourages the model to actively rely on tools during reasoning, rather than falling back to pure internal reasoning.

More importantly, although the tool-call success rate of \iapo\ is initially low, it rises sharply in the later stage of training and eventually approaches 100\%, significantly surpassing VTool-R1. We hypothesize that this delayed improvement arises because \iapo\ optimizes a stricter objective: the model must not only call a tool, but also ground that decision in the appropriate input evidence. As a result, the model first learns to increase its willingness to use tools, reflected in the high tool-call ratio, and then gradually aligns its input attribution with the teacher’s guidance, which leads to a rapid improvement in tool-selection accuracy. In contrast, VTool-R1 lacks such process-level supervision, resulting in both a lower tool-call ratio and a relatively stagnant tool-call success rate. 

\section{Conclusion}
In this paper, we propose \iapo\, a method that improves multimodal SLM tool use through attribution-guided policy optimization. We introduce an integrated gradients–based attribution penalty to quantify how different parts of the input contribute to the generated tool call, and use the discrepancy between the student’s IG scores and those of an external teacher as a penalty during training. This attribution-based guidance encourages the model to attend to the most relevant input regions, leading to better tool selection. Through case studies and experiments on a widely used visual tool-calling framework, we demonstrate both the soundness of our motivation and the effectiveness of \iapo\ method.

\bibliography{colm2026_conference}
\bibliographystyle{colm2026_conference}

\clearpage
\appendix
\section{Dataset} \label{app:dataset}
Our training and evaluation data are adapted from \citet{fu2025refocus}. We consider two categories of tasks, \textbf{Chart} and \textbf{Table}, both of which require structured visual reasoning over charts or tables.
\begin{itemize}[leftmargin=*, nosep]
    \item The \textbf{Chart} tasks include data from CharXiv \citep{wang2024charxiv} and ChartQA \citep{masry2022chartqa}. The ChartQA dataset provides 14,344 training examples and 813 validation examples, covering a mixture of vertical and horizontal bar charts. In addition, ReFocus collects 444 VQA pairs on horizontal bar charts and 382 VQA pairs on vertical bar charts for testing. The CharXiv dataset is used only for evaluation, serving as a benchmark for assessing the generation ability of different methods, and contains 143 visual question answering (VQA) pairs.
    \item The \textbf{Table} tasks include three datasets: VWTQ \citep{pasupat2015wikitablequestions}, VWTQ\_syn \citep{kim2024tablevqabench}, and VTabFact \citep{chen2020tabfact}. For each dataset, we use 70\% of samples for training and reserve the remaining 30\% for evaluation, yielding three separate test sets.
\end{itemize}
\section{Ablation} \label{sec:ablation}

\subsection{Ablation for Teacher models}
In this work, we consider two types of teacher models: off-the-shelf models used directly as teachers, and RL-trained models used as teachers. We further investigate whether a more advanced model can provide stronger supervision. Table~\ref{tab:techer_ab} reports the validation accuracy at 200 training steps for different teacher choices. We observe that the RL-trained teacher achieves the best performance. This result is consistent with Table~\ref{tab:baselines}, where the RL-trained 7B model performs substantially better than the RL-trained 3B model, indicating that it can provide more effective guidance for small-model tool calling. In contrast, using a more advanced teacher such as Qwen3-VL-8B \citep{bai2025qwen3} does not lead to optimal performance, likely because the larger discrepancy in internal mechanisms makes its attribution patterns less suitable for transferring to the student. Therefore, we recommend selecting teachers from the same model family whenever possible.
\begin{table}[h]
\centering
\small
\begin{tabular}{ccccc}
\toprule
Model    & Qwen-2.5VL-7B & 7B+GRPO & Qwen3-VL-8B & 8B+GRPO \\
Accuracy & 68.5          & 70.4    & 64.3        & 65.2    \\ \bottomrule
\end{tabular}
\caption{Ablation for teacher models.}
\label{tab:techer_ab}
\end{table}

\subsection{Ablation for Hyper-parameter search}
We also conduct a hyperparameter study on the importance-score coefficient $\lambda$ and used in IAPO. The results are reported in Table~\ref{tab:lambda_ab}. As shown in the table, a relatively small coefficient of 0.1 combined with a decay rate of 0.01 achieves the best performance. 

\begin{table}[h]
\centering
\small
\begin{tabular}{ccccc}
\toprule
$\lambda$    & 0.05 & 0.1  & 0.5  & 1    \\
Accuracy & 68.1 & 70.4 & 69.4 & 64.4 \\ \bottomrule
\end{tabular}
\caption{Ablation for importance-score coefficient $\lambda$.}
\label{tab:lambda_ab}
\end{table}


\section{Full Prompt and Prompt Decomposition Strategy} \label{app:prompt}

\begin{tcolorbox}[
    halign=flush left,
    breakable, 
    colback=teal!5!white,
    colframe=teal!75!black,
    title={\small\textbf{System prompt template with an example division}},
    fonttitle=\bfseries,
    fontupper=\ttfamily\footnotesize,
    width=\columnwidth,  
    enhanced jigsaw,
]
{\color{red} \{Image $I$\}} {\color{blue} <Image tokens>} {\color{blue} </Image tokens>}

{\color{red} \{User Query $R$\}} {\color{blue} <Query>} {\color{blue} </Query>}

{\color{red} \{Pure-text $P_1$\}}  Here are some tools that can help you. All are python codes. They are in tools.py and will be imported for you.
You will be given a table figure: image\_1 and a question.
Notice that you, as an AI assistant, are not good at answering questions when there are too many unnecessary and irrelevant information. You should determine which are the relevant columns to the question, and specify them in a python list. You should use the given column headers.
You should also determine which are the relevant rows to the question, and specify them in a python list. You should use the given row headers.
You could select the tools to focus on some columns / rows, or mask out some columns / rows. Use whichever tool you think is more appropriate.
Below are the tools in tools.py:
{\color{gray}\begin{lstlisting}[keepspaces=true, basicstyle=\ttfamily, breaklines=True, escapeinside={(*@}{@*)}]
```python
(*@{\color{red}\{Tool Definition $T_1$\}}@*)
def focus_on_columns_with_highlight(image, columns_to_focus_on, all_columns_bounding_boxes):
    \"\"\"
    This function is useful when you want to focus on some specific columns of the image.
    It does this by adding light transparent red highlight to the columns that need to be focused on.
    For example, you can focus on the columns in a table that are relevant to your analysis.
    Return the drawed image.

    Args:
        image (PIL.Image.Image): the input image
        columns_to_mask (List[str]): a list of column names to focus on.
        all_columns_bounding_boxes (Dict[Dict]]): a dictionary of bounding boxes for all columns in the image. key is column name and value is the bounding box of that column. Each bounding box is in the format {'x1': x1, 'y1': y1, 'x2': x2, 'y2': y2}.

    Returns:
        image_with_focused_columns (PIL.Image.Image): the image with specified columns focused on
        
    Example:
        image = Image.open("sample_img.jpg")
        image_with_focused_columns = focus_on_columns_with_highlight(image, ["Year", "Name"], {"Year": {'x1': 0.1, 'y1': 0.1, 'x2': 0.3, 'y2': 0.9}, "Team": {'x1': 0.4, 'y1': 0.1, 'x2': 0.6, 'y2': 0.9}, "Name": {'x1': 0.7, 'y1': 0.1, 'x2': 0.9, 'y2': 0.9}})
        display(image_with_focused_columns)
    \"\"\"
(*@{\color{red}\{Tool Definition $T_2$\}}@*)
def focus_on_rows_with_highlight(image, rows_to_focus_on, all_rows_bounding_boxes):
    \"\"\"
    This function is useful when you want to focus on some specific rows of the image.
    It does this by adding light transparent red highlight to the rows that need to be focused on.
    For example, you can focus on the rows in a table that are relevant to your analysis.
    Return the drawed image.
    
    Args:
        image (PIL.Image.Image): the input image
        rows_to_focus_on (List[str]): a list of row headers to focus on.
        all_rows_bounding_boxes (Dict[Dict]): a dictionary of bounding boxes for all rows in the image. key is row header and value is the bounding box of that row. Each bounding box is in the format {'x1': x1, 'y1': y1, 'x2': x2, 'y2': y2}.
    
    Returns:
        image_with_focused_rows (PIL.Image.Image): the image with specified rows focused on

    Example:
        image = Image.open("sample_img.jpg")
        image_with_focused_rows = focus_on_rows_with_highlight(image, ["1972"], ["Year": {'x1': 0.1, 'y1': 0.1, 'x2': 0.9, 'y2': 0.15}, "1969": {'x1': 0.1, 'y1': 0.2, 'x2': 0.9, 'y2': 0.5}, "1972": {'x1': 0.1, 'y1': 0.6, 'x2': 0.9, 'y2': 0.9}])
        display(image_with_focused_rows)
    \"\"\"
(*@{\color{red}\{Tool Definition $T_3$\}}@*)
def focus_on_columns_with_mask(image, columns_to_focus_on, all_columns_bounding_boxes):
    \"\"\"
    This function is useful when you want to focus on some specific columns of the image.
    It does this by masking out the columns that are not needed.
    For example, you can focus on the columns in a table that are relevant to your analysis and ignore the rest.
    Return the masked image.

    Args:
        image (PIL.Image.Image): the input image
        columns_to_mask (List[str]): a list of column names to focus on.
        all_columns_bounding_boxes (Dict[Dict]]): a dictionary of bounding boxes for all columns in the image. key is column name and value is the bounding box of that column. Each bounding box is in the format {'x1': x1, 'y1': y1, 'x2': x2, 'y2': y2}.

    Returns:
        image_with_focused_columns (PIL.Image.Image): the image with specified columns focused on
        
    Example:
        image = Image.open("sample_img.jpg")
        image_with_focused_columns = focus_on_columns(image, ["Year", "Name"], {"Year": {'x1': 0.1, 'y1': 0.1, 'x2': 0.3, 'y2': 0.9}, "Team": {'x1': 0.4, 'y1': 0.1, 'x2': 0.6, 'y2': 0.9}, "Name": {'x1': 0.7, 'y1': 0.1, 'x2': 0.9, 'y2': 0.9}})
        display(image_with_focused_columns)
    \"\"\"
(*@{\color{red}\{Tool Definition $T_4$\}}@*)
def focus_on_rows_with_mask(image, rows_to_focus_on, all_rows_bounding_boxes):
    \"\"\"
    This function is useful when you want to focus on some specific rows of the image.
    It does this by masking out the rows that are not needed.
    For example, you can focus on the rows in a table that are relevant to your analysis and ignore the rest.
    Return the masked image.
    
    Args:
        image (PIL.Image.Image): the input image
        rows_to_focus_on (List[str]): a list of row headers to focus on.
        all_rows_bounding_boxes (Dict[Dict]): a dictionary of bounding boxes for all rows in the image. key is row header and value is the bounding box of that row. Each bounding box is in the format {'x1': x1, 'y1': y1, 'x2': x2, 'y2': y2}.
    
    Returns:
        image_with_focused_rows (PIL.Image.Image): the image with specified rows focused on

    Example:
        image = Image.open("sample_img.jpg")
        image_with_focused_rows = focus_on_rows(image, ["1972"], ["Year": {'x1': 0.1, 'y1': 0.1, 'x2': 0.9, 'y2': 0.15}, "1969": {'x1': 0.1, 'y1': 0.2, 'x2': 0.9, 'y2': 0.5}, "1972": {'x1': 0.1, 'y1': 0.6, 'x2': 0.9, 'y2': 0.9}])
        display(image_with_focused_rows)
    \"\"\"
(*@{\color{red}\{Tool Definition $T_5$\}}@*)
def focus_on_columns_with_draw(image, columns_to_focus_on, all_columns_bounding_boxes):
    \"\"\"
    This function is useful when you want to focus on some specific columns of the image.
    It does this by drawing a red box around the columns that need to be focused on.
    For example, you can focus on the columns in a table that are relevant to your analysis.
    Return the drawed image.

    Args:
        image (PIL.Image.Image): the input image
        columns_to_mask (List[str]): a list of column names to focus on.
        all_columns_bounding_boxes (Dict[Dict]]): a dictionary of bounding boxes for all columns in the image. key is column name and value is the bounding box of that column. Each bounding box is in the format {'x1': x1, 'y1': y1, 'x2': x2, 'y2': y2}.

    Returns:
        image_with_focused_columns (PIL.Image.Image): the image with specified columns focused on
        
    Example:
        image = Image.open("sample_img.jpg")
        image_with_focused_columns = focus_on_columns(image, ["Year", "Name"], {"Year": {'x1': 0.1, 'y1': 0.1, 'x2': 0.3, 'y2': 0.9}, "Team": {'x1': 0.4, 'y1': 0.1, 'x2': 0.6, 'y2': 0.9}, "Name": {'x1': 0.7, 'y1': 0.1, 'x2': 0.9, 'y2': 0.9}})
        display(image_with_focused_columns)
    \"\"\"
(*@{\color{red}\{Tool Definition $T_6$\}}@*)
def focus_on_rows_with_draw(image, rows_to_focus_on, all_rows_bounding_boxes):
    \"\"\"
    This function is useful when you want to focus on some specific rows of the image.
    It does this by drawing a red box around the rows that need to be focused on.
    For example, you can focus on the rows in a table that are relevant to your analysis.
    Return the drawed image.
    
    Args:
        image (PIL.Image.Image): the input image
        rows_to_focus_on (List[str]): a list of row headers to focus on.
        all_rows_bounding_boxes (Dict[Dict]): a dictionary of bounding boxes for all rows in the image. key is row header and value is the bounding box of that row. Each bounding box is in the format {'x1': x1, 'y1': y1, 'x2': x2, 'y2': y2}.
    
    Returns:
        image_with_focused_rows (PIL.Image.Image): the image with specified rows focused on

    Example:
        image = Image.open("sample_img.jpg")
        image_with_focused_rows = focus_on_columns_with_highlight(image, ["1972"], ["Year": {'x1': 0.1, 'y1': 0.1, 'x2': 0.9, 'y2': 0.15}, "1969": {'x1': 0.1, 'y1': 0.2, 'x2': 0.9, 'y2': 0.5}, "1972": {'x1': 0.1, 'y1': 0.6, 'x2': 0.9, 'y2': 0.9}])
        display(image_with_focused_rows)
    \"\"\"
```\end{lstlisting}}
{\color{red} \{Pure-text $P_2$\}} 
{\color{orange}\# GOAL \#:} Based on the above tools, I want you to reason about how to solve the {\color{orange}\# USER REQUEST \#} and generate the actions step by step (each action is a python function call) to solve the request.
You may need to use the tools above to process the images and make decisions based on the visual outputs of the previous code blocks.
You should only use the tools above, you should not use other functions or code which will not be executed.

{\color{orange}\# REQUIREMENTS \#:}

1. The generated actions can resolve the given user request {\color{orange}\# USER REQUEST \#} perfectly. The user request is reasonable and can be solved. Try your best to solve the request.

2. The arguments of a tool must be the same format specified in {\color{orange}\# TOOL LIST \#};

3. If you think you got the answer, use {\color{brown}ANSWER:} {\color{blue}<your answer>} Please extract the final answer in {\color{brown}FINAL ANSWER:} {\color{blue}<final answer>} and ends with {\color{brown}TERMINATE}.

4. All images in the initial user request are stored in PIL Image objects named image\_1, image\_2, ..., image\_n. You can use these images in your code blocks. Use display() function to show the image in the notebook for you too see.

5. Use as few tools as possible. Only use the tools for the use cases written in the tool description. You can use multiple tools in a single action.

6. If you have multiple answers, please separate them with || marks. For example, if the answer is 'Alice' and 'Bob', you should write 'Alice||Bob'.

7. When you focus on columns in the image, most like you need to look at multiple columns instead of a single one. 

8. If you do not think you have enough information to answer the question on the images returned by the tools, you should directly answer the question based on the original image.

Below are some examples of how to use the tools to solve the user requests. You can refer to them for help. You can also refer to the tool descriptions for more information.

9. Only one turn of action, {\color{brown}ACTION 0}, is allowed. You must provide the answer after maximum one {\color{brown}ACTION} call.

\vspace{1em}
{\color{orange}\# EXAMPLE:}  Simple question that does not require any tool

{\color{orange}\# USER REQUEST \#:} {\color{blue} <A image here>} What is the title of this table?

{\color{orange}\# USER Bounding Box Info:} columns\_bbox, where keys are column headers and values are column bounding boxes. rows\_bbox, where keys are row headers and values are row bounding boxes. The columns in the image are: ["Grade", "Mentor", "Salary"]. The rows in the image start with: ["Grade", "A", "B", "C"].

{\color{orange}\# USER IMAGE} stored in image\_1, as PIL image.

{\color{orange}\# RESULT \#:}

{\color{brown}THOUGHT 0:} The question does not require any tool. I can see the title of the table is "Customer Information".

{\color{brown}ACTION 0:} No action needed.

{\color{brown}ANSWER:} The title of the table is "Customer Information". {\color{brown}FINAL ANSWER:} Customer Information. {\color{brown}TERMINATE}

\vspace{1em}
{\color{orange}\# EXAMPLE:}  Focus on specific columns in the image

{\color{orange}\# USER REQUEST \#:} {\color{blue} <A image here>} Who had the same game version as John Roth?

{\color{orange}\# USER Bounding Box Info:} columns\_bbox, where keys are column headers and values are column bounding boxes. rows\_bbox, where keys are row headers and values are row bounding boxes. The columns in the image are: ['Manager Name', 'Game Version', 'Game Score']. The rows in the image start with: ['Manager Name', 'John Roth', 'Alice Smith', 'Bob Johnson'].

{\color{orange}\# USER IMAGE} stored in image\_1, as PIL image.

{\color{orange}\# RESULT \#:}

{\color{brown}THOUGHT 0:} To identify who had the same game version as John Roth, I need to focus on the 'Game Version' column and the 'Manager Name' column. Also, I need to focus on all the rows so I do not need to focus on some specific rows.

{\color{brown}ACTION 0:}
{\color{gray}
\begin{lstlisting}[keepspaces=true, basicstyle=\ttfamily, breaklines=True]
```python
image_with_focused_columns = focus_on_columns_with_draw(image\_1, ["Game Version", "Manager Name"], columns\_bbox)
display(image_with_focused_columns)
```
\end{lstlisting}}
{\color{brown}OBSERVATION:} Execution success. The output is as follows:

{\color{blue} <the image outputs of the code is added as the second image>}

{\color{brown}THOUGHT 1:} Now I can see the 'Game Version' column and the 'Manager Name' column more clearly. The game version of John Roth is 'v1.2'. Other people with the same game version are 'Alice Smith' and 'Bob Johnson'.

{\color{brown}ANSWER:} 'Alice Smith' and 'Bob Johnson' had the same game version as John Roth are. {\color{brown}FINAL ANSWER:} Alice Smith||Bob Johnson. {\color{brown}TERMINATE}

\vspace{1em}
{\color{orange}\# EXAMPLE:}  Focus on specific rows in the image

{\color{orange}\# USER REQUEST \#:} {\color{blue} <A image here>} How many games did Josh win after 1996? 

{\color{orange}\# USER Bounding Box Info:} columns\_bbox, where keys are column names and values are column bounding boxes. rows\_bbox, where keys are row headers and values are row bounding boxes. The columns in the image are: ["Rank", "Year", "Score", "Month"]. The rows in the image start with: ["Rank", "0", "1", "2", "3", "4", "5"].

{\color{orange}\# USER IMAGE} stored in image\_1, as PIL image.

{\color{orange}\# RESULT \#:}

{\color{brown}THOUGHT 0:} This table is about the games played by Josh that he won. I do not need to focus on any specific columns. I need to focus on the rows with the year after 1996. Three rows have year after 1996: one has year 1997 and this row starts with "3", one has year 1998 and this row starts with "4", and one has year 1999 and this row starts with "5". So I will focus on the rows with row starters "3", "4", and "5".

{\color{brown}ACTION 0:}
{\color{gray}
\begin{lstlisting}[keepspaces=true, basicstyle=\ttfamily, breaklines=True]
```python
image_with_focused_rows = focus_on_rows_with_highlight(image\_1, ["3", "4", "5"], rows\_bbox)
display(image_with_focused_rows)
```
\end{lstlisting}}
{\color{brown}OBSERVATION:} Execution success. The output is as follows:

{\color{blue} <the image outputs of the code is added as the second image>}

{\color{brown}THOUGHT 1:} I can see that Josh won 3 games after 1996.

{\color{brown}ANSWER:} Josh won 3 games after 1996. {\color{brown}FINAL ANSWER:} 3. {\color{brown}TERMINATE}

\vspace{1em}
{\color{orange}\# EXAMPLE:}  Focus on specific columns and specific rows in the image

{\color{orange}\# USER REQUEST \#:} {\color{blue} <A image here>} what is the sum of annual earnings after 2006? 

{\color{orange}\# USER Bounding Box Info:} columns\_bbox where keys are column names and values are column bounding boxes. rows\_bbox, where keys are row headers and values are row bounding boxes. The columns in the image are: ["Index", "Year", "Cost", "Earning"]. The rows in the image start with: ["Index", "0", "1", "2", "3", "4", "5"].

{\color{orange}\# USER IMAGE} stored in image\_1, as PIL image.

{\color{orange}\# RESULT \#:}

{\color{brown}THOUGHT 0:} I need to focus on the 'Year' column and the 'Earning' column. I also need to focus on the rows with years after 2006. The row with year being 2006 starts with "3". So I will focus on the rows starting with "4", and "5".

{\color{brown}ACTION 0:}
{\color{gray}
\begin{lstlisting}[keepspaces=true, basicstyle=\ttfamily, breaklines=True]
```python
image_with_focused_columns = focus_on_columns_with_mask(image\_1, ["Year", "Earning"], columns\_bbox)
image_with_focused_rows = focus_on_rows_with_draw(image_with_focused_columns, ["4", "5"], rows\_bbox)
display(image_with_focused_rows)
```
\end{lstlisting}}
{\color{brown}OBSERVATION:} Execution success. The output is as follows:

{\color{blue} <the image outputs of the code is added as the second image>}

{\color{brown}THOUGHT 1:} I can see that the annual earnings after 2006 are \$165,498 and \$198,765. The sum of the annual earnings after 2006 is \$364,263.

{\color{brown}ANSWER:} The sum of the annual earnings after 2006 is \$364,263. {\color{brown}FINAL ANSWER:} 364263. {\color{brown}TERMINATE}.

{\color{orange}\# USER Bounding Box Info:} x\_values\_bbox, storing x values and coordinates. y\_values\_bbox, storing x values and coordinates. The x values in the image are:  {\color{blue}<x\_values>}. The y values in the image are:  {\color{blue}<y\_values>}.

{\color{orange}\# USER IMAGE} stored in image\_1, as PIL image. 
\end{tcolorbox}

\end{document}